\documentclass[conference]{IEEEtran}
\IEEEoverridecommandlockouts
\usepackage{cite}
\usepackage{amsmath,amssymb,amsfonts}
\usepackage{algorithmic}
\usepackage{graphicx}
\usepackage{textcomp}
\usepackage{xcolor}
\usepackage{hyperref}
\def\BibTeX{{\rm B\kern-.05em{\sc i\kern-.025em b}\kern-.08em
    T\kern-.1667em\lower.7ex\hbox{E}\kern-.125emX}}
\begin{document}

\title{Enhancing Satellite Imagery using Deep Learning for the Sensor To Shooter Timeline}

\author{\IEEEauthorblockN{Matthew Ciolino}
\IEEEauthorblockA{\textit{PeopleTec Inc.} \\
Huntsville, AL, USA \\
matt.ciolino@peopletec.com}
\and
\IEEEauthorblockN{Dominick Hambrick}
\IEEEauthorblockA{\textit{PeopleTec Inc.} \\
Huntsville, AL, USA \\
dominick.hambrick@peopletec.com}
\and
\IEEEauthorblockN{David Noever}
\IEEEauthorblockA{\textit{PeopleTec Inc.} \\
Huntsville, AL, USA \\
david.noever@peopletec.com}
}

\maketitle

\begin{abstract}
The sensor to shooter timeline is affected by two main variables: satellite positioning and asset positioning. Speeding up satellite positioning by adding more sensors or by decreasing processing time is important only if there is a prepared shooter, otherwise the main source of time is getting the shooter into position. However, the intelligence community should work towards the exploitation of sensors to the highest speed and effectiveness possible. Achieving a high effectiveness while keeping speed high is a tradeoff that must be considered in the sensor to shooter timeline. In this paper we investigate two main ideas, increasing the effectiveness of satellite imagery through image manipulation and how on-board image manipulation would affect the sensor to shooter timeline. We cover these ideas in four scenarios: Discrete Event Simulation of onboard processing versus ground station processing, quality of information with cloud cover removal, information improvement with super resolution, and data reduction with image to caption. This paper will show how image manipulation techniques such as Super Resolution, Cloud Removal, and Image to Caption will improve the quality of delivered information in addition to showing how those processes effect the sensor to shooter timeline.
\end{abstract}


\section{Introduction}
In this paper we investigate improvements in the speed of delivery and improvements in the quality of information of the satellite imagery process. This can help be explained by imagining a scenario or battle network \cite{battle} in which our timeline will operate. Think about the following battle network: ally units are advancing over a hillside in which they suspect enemy units occupy. A request is made to take a picture over the hill to assess any threats. This request goes to a ground station, is approved, or automatically sent as a task to a satellite. The satellite waits to get into position and takes a picture of the location of interest. The satellite then waits to get into view of a ground station or sends the information along a relay. The image is then sent from the ground station to the ally unit which the unit then views.

While this simple scenario in real life is more of a densely connected web of connections with data relays, ground stations, and various satellite configurations, it is easy to understand where problems might arise. There could be limited connection in the ally unit’s area to send/receive data, there could be a lack of understanding of what might be in the image (radar jammer vs anti air), there could be a cloud in the image which obscures a target of interest, there could be a lack of resolution in the image so a smaller target could not be identified, there could be a lack of physical assets to respond to a newfound threat. All these scenarios might arise and therefore must be considered to better improve the system.

Both the US/UK Interoperability Study \cite{us/uk} and the Australian DoD’s Joint Operations and Analysis Division \cite{new} consider the “man in the loop” time to analyze an image and the positioning of physical assets as the biggest factors in improving the sensor to shooter timeline. This need for increased capabilities is corroborated by the Congressional Research Service \cite{intelligence}, stating that growth in remote sensing for both hardware and software is paramount for future of the Intelligence, Surveillance, and Reconnaissance (ISR) community. This paper will focus on how improving the “man in the loop” process impacts the satellite sensor to shooter timeline.

\subsection{Background}
Discrete Event Simulation (DES) is a method used to model real world systems that can be decomposed into a set of logically separate processes that autonomously progress through time. Each event occurs on a specific process and is assigned a timestamp \cite{discrete}. Commonly used to understand factors that impact waiting times at various entry points (concert, hospital), can be adapted to work for a simple satellite constellation simulation. Various simulation tools exist for modeling satellite communications, power, and attitude control \cite{freimann2021estnet} \cite{wood2012savi}.

Super Resolution (SR) is the process of taking a low-resolution image and running it through Machine Learning (ML) model to increase the resolution with higher fidelity information than any traditional scaling algorithm \cite{yang2019deep} \cite{wang2020deep}. SR, when used as a preprocessing technique, can increase the performance of downstream computer vision tasks by up to 15\% \cite{ciolino2020training} \cite{shermeyer2019effects}. SR has recently been tested in a technology demonstration for onboard satellite super resolution by Lockheed Martin via a Jetson Nano \cite{lockheed_martin_2022}.

Total cloud cover is 58\% on land and 72\% over oceans \cite{warren_eastman_hahn_2014}. Visible spectrum satellites which provide the highest definition imagery in color are unable to see the ground most of the time. While this makes a great case for other sensors (Synthetic Aperture Radar (SAR), Infrared (IR)), it severely limits the most popular form of satellite imagery \cite{visible}, visible. Of the over 50 countries \cite{toth2016remote} operating over 589 remote sensing satellites (labeled imaging) \cite{union}, 76\% (448) of them are for optical imaging. This leads to a huge need in the community to both avoid taking pictures of a cloud area, avoiding downloading clouds from the satellite, and interestingly, attempt to remove the clouds overlaying the location of interest. Many researchers have attempted cloud removal \cite{lin2019remote} \cite{zhao2021seeing} \cite{chen2019thick} \cite{darbaghshahi2021cloud} and cloud detection \cite{shan2009onboard} \cite{zhang2019cloud} \cite{li2019deep} using small machine learning models.

In bandwidth constrained locations, sending an entire image to evaluate can be impossible. In such a scenario using a way to dilute an entire image into a couple of sentences or summary becomes advantageous. This is the goal of an image to caption model \cite{noever2020discoverability} \cite{vinyals2015show} \cite{herdade2019image}. These machine learning models combine a computer vision (CV) model with a natural language processing (NLP) model to compress an image into a text description. For example, in the RSCID dataset \cite{lu2017exploring}, airplanes on the tarmac might be labeled “Four planes are parked next to two buildings on an airport.”. Using this system in the example battle network in the introduction above might yield the count of enemy units in an area in text format.

\subsection{Contributions}
For the speed of delivery, we look at the process of requesting and image, sending that request to a satellite, taking a picture, analyzing the image, and sending that information back to the requestee.  This is the sensor to shooter timeline which we will dissect in this paper. Using a DES, we will modulate different parts of this timeline to see how they impact the overall duration. The first scenario considered is onboard satellite versus traditional ground station processing. This will show how changing where the image is analyzed impacts the timeline. The second scenario attempts to capture a worst-case scenario in a satellite constellation, only access to one sensor or one ground station. The third scenario is showing how increased processing time (i.e., increased quality of information) would impact the timeline.

For the quality of information, we look at improving the information received from a satellite image mainly through the application of ML.  In the first scenario, we apply SR, the process of increasing the size of an image, to images and see how downstream tasks are improved. In the second scenario, we see how even a small number of clouds would impact downstream ML tasks. In the third scenario, we show how an image to caption model might be advantageous in a low connectivity environment. The impact on the sensor to shooter timeline is talked about in the speed of delivery section while the improvement to information is talked about in the obscured images, enhanced images, and the improved response sections.

\section{Experiments}

\subsection{Speed of Delivery}

For each of these scenarios (onboard vs ground station processing, worst case single satellite/ground station, and increased processing time) we describe a base case [Table \ref{tab:discrete-event-simulation}] which we modify our discrete event simulation to emulate the impact of these changes. From a base case, we run the three scenarios: ground station processing, onboard processing, and increased processing time. For each of these three scenarios we run three simulations: all satellites and ground stations, one satellite in use, and one ground station in use. We then modulate the number of images, the number of satellites, and the speed of processing for a total of 36 simulations.

\subsubsection{Discrete Event Simulation}

Using SimPy \cite{overview}, a process-based DES tool, we can construct a real-time simulation using time-based events. We work with this tool through the work of gate-simulation \cite{dattivo}, a DES that incorporates a two-stage linked approach between requests and providers. This allows us to extend this work for use for a satellite simulation \cite{ciolino}. We operate a theoretical system of a 3-satellite constellation with 5 ground station links.  Each simulation runs based on the following base configuration. We modify these values for each variation.

\begin{table}[]
\centering
\caption{Discrete Event Simulation Base Case}
\label{tab:discrete-event-simulation}
\begin{tabular}{|c|c|c|}
\hline
\textbf{Event} & \textbf{Value} & \textbf{Units} \\ \hline
Img Request Interval & 30 & minutes \\ \hline
Mean Num Images Requests & 20 & images \\ \hline
Std Num Images Requests & 5 & images \\ \hline
Mean Data Transfer to Sat & 30 & seconds \\ \hline
Std Data Transfer to Sat & 6 & seconds \\ \hline
Mean Downlink to Shooter & 6 & seconds \\ \hline
Std Downlink to Shooter & 0.6 & seconds \\ \hline
Num Sats & 3 & satellites \\ \hline
Sat Time to Picture & 4* & minutes \\ \hline
Num Ground Station & 5 & ground stations \\ \hline
Time to Ground Station Link & 3* & minutes \\ \hline
Ground Station Processing & 3 & seconds \\ \hline
Onboard Sat to Shooter & 6 & seconds \\ \hline
Onboard Sat Processing & 6 & seconds \\ \hline
Onboard Sat SR & 6 & seconds \\ \hline
\end{tabular} \\
*represents an exponential function: $f(x) = exp^{-x}$
\end{table}

\subsection{Quality of Information}

For each of the scenarios from the battle network in the introduction we show a developed solution. We show the experiment used for cloud detection, super resolution, and image to caption. 

\subsubsection{Obscured Images}

\begin{figure*}[!t]
    \centering
    \includegraphics[width=\textwidth]{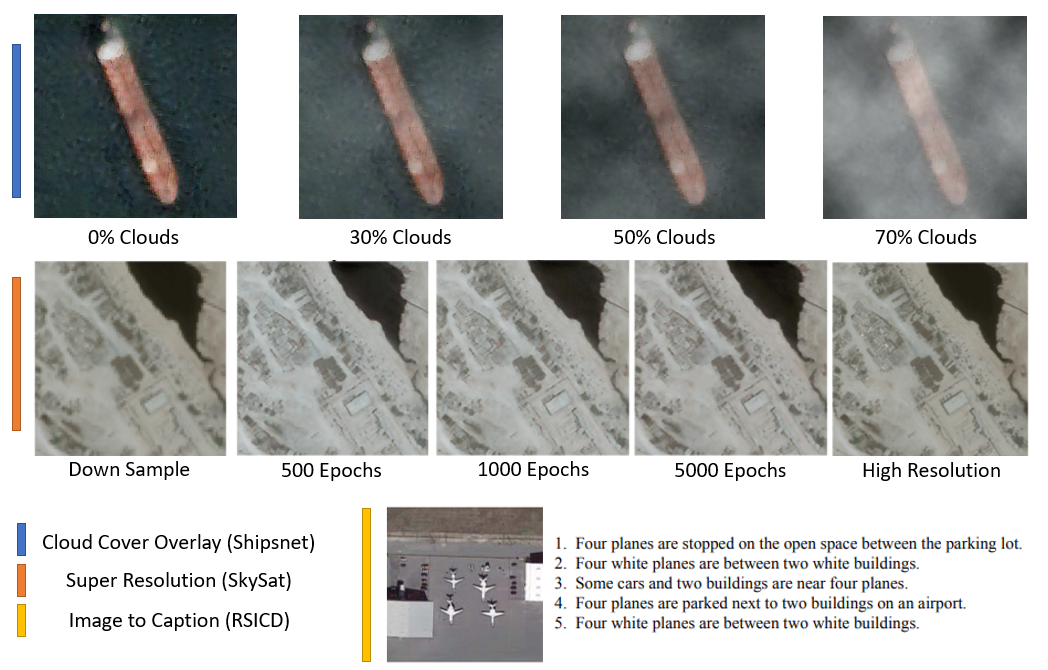}
    \caption{Datasets and Examples of Cloud Cover (Blue), Super Resolution (Orange), Image to Caption (Yellow)}
    \label{data}
\end{figure*}

The presence of clouds, even partially cloudy images, might significantly impact the performance of a machine learning algorithm. Similar to the RICE \cite{lin2019remote} dataset, we took the Shipsnet \cite{rhammell_2018} dataset, image chips extracted from Planet satellite imagery collected over the San Francisco Bay and San Pedro Bay areas of California, and overlayed an image of a cloud with an alpha layer [Figure \ref{data}]. This allowed us to control the opacity of the clouds to simulate partial cloud cover versus total cloud cover. We then trained a small neural network to run a binary classifier to predict if there was a ship or not ship in the image. After training (4-ConV, 2 Dense) to an almost 95\% accuracy, we then ran inference on each cloud composite from 10\% to 100\% clouds to inspect the loss in accuracy. 

\subsubsection{Enhanced Images}

A high-resolution dataset is needed for an effective SR model to be trained. We compiled Skysat’s 0.8m samples \cite{rousmaniere} totaling over 1.5 GB of pansharpened, orthorectified, color corrected RGB images. To train a model to up sample an image to a desired resolution we must first down sample the high-resolution dataset using a scaling algorithm such as bicubic. While state of the art has moved to a machine learning based down sampler \cite{sun2020learned}, traditional down scaling works well. We then use the SRGAN \cite{ciolino2020training} \cite{wang2020deep} network to up sample the low-resolution image back into the original high resolution. We can see the improvement in up sample during the course of training [Figure \ref{data}]. To see how super resolving impacts downstream tasks, we super resolved the Shipsnet dataset \cite{rhammell_2018} and compared the performance of the SR images and the raw images on the MaskRCNN \cite{he2017mask} model using mean average precision (mAP). 

\subsubsection{Improved Response}

Image to caption models work by combining a CV model and an NLP model. The CV model extracts a feature embedding which is then used in an NLP model to predict a text output. We use the RSICD dataset \cite{lu2017exploring} which is composed of 10,921 satellite images with 50,000 text descriptions at a 70/30 train test split [Figure \ref{data}]. We trained various CV models (VGG16/19 \cite{simonyan2014very}, NASNetMobile \cite{zoph2018learning}) as a backbone to see the difference in the text output. We run inference on random xView images \cite{lam2018xview}, a collection of 1 million objects in satellite imagery. We evaluate performance with the BLEU \cite{papineni2002bleu} score comparing a text prediction with the labels.

\section{Evaluation}

\subsection{Discrete Event Simulation}

\begin{figure*}[!t]
    \centering
    \includegraphics[width=\textwidth]{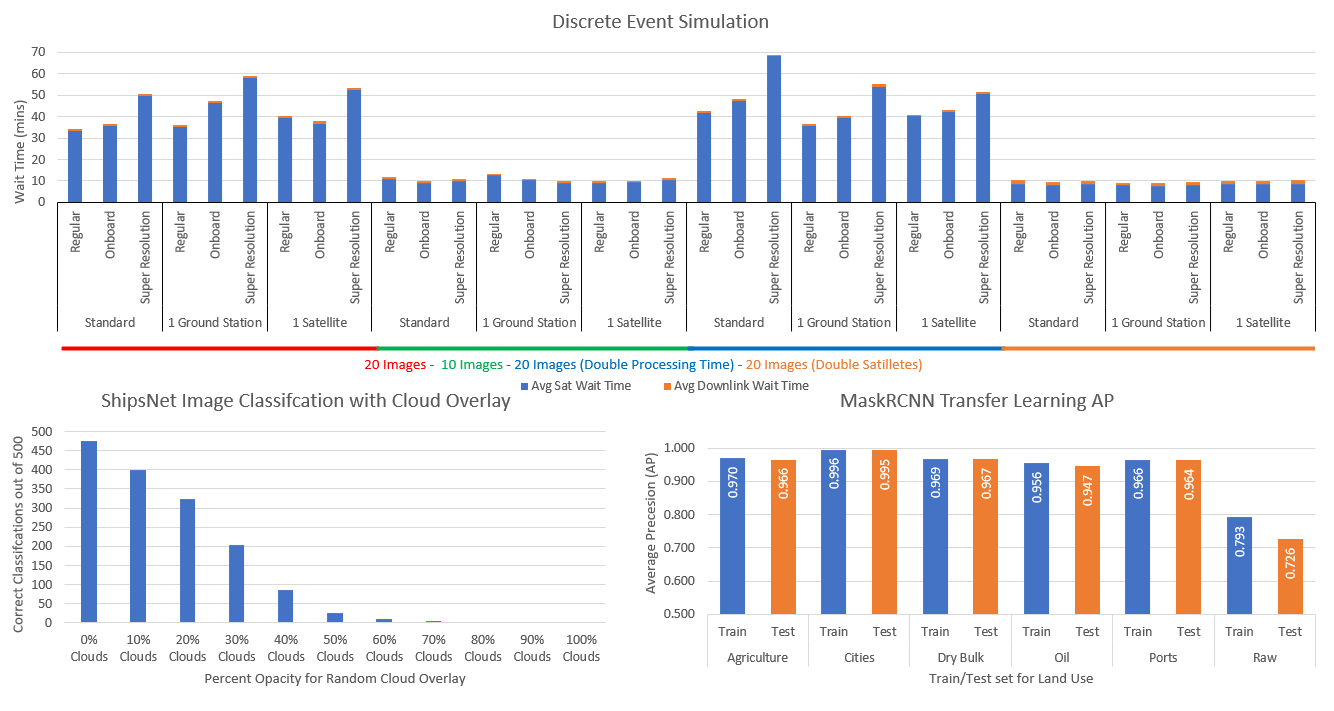}
    \caption{Discrete Event Simulation Results, Shipsnet Cloud Overlay Results, and Super Resolution AP Results}
    \label{results}
\end{figure*}

Our DES was able to show results across 36 simulations [Figure \ref{results}] with both the average wait time to send an image request to a satellite (blue) and the average wait time to send an image back down (orange). For our base case of around 20 images requested every 30 mins, we found an average total wait time of 34.4 minutes. This increase when we limited the simulation to onboard compute (36.5m, +6.10\%) and added SR (50.6m, +47.09\%). We also saw increases when we limited the simulation to 1 ground station (36.1m, +4.94\%) and 1 satellite (40.3m, +17.15\%). The base case was also altered for half the number of images (11.7m, -65.99\%), double the processing time (42.5m, +23.55\%), and double the number of satellites (10.6m, -69.19\%). Overall, we found that the biggest factor in the timeline is waiting for the satellite to get into the correct location to take a picture with only 3\% to 20\% of the time waiting for ground station processing.

\subsection{Obscured Images}

As shown, cloud cover significantly impacts the ability for a normally trained (non-generalizable) network to perform when even a small amount of reflections from clouds obscure the image [Figure \ref{results}]. In terms of accuracy, the model performed 79.8\% at 10\% clouds, 65.0\% at 20\% clouds, 40.8\% at 30\% clouds, 17.4\% at 40\% clouds, and less than 5\% to 0\% when there is medium to total cloud coverage in an area. The need for cloud detection and removal is paramount to avoid processing unusable pictures and finite downlink bandwidth.

\subsection{Enhanced Images}

Super resolution across many different land use cases with various image compositions is shown to increase downstream object detection performance by 27.66\% (75.95\% to 96.96\%) [Figure \ref{results}].

\subsection{Improved Response}

\begin{figure}[t]
    \centering
\includegraphics[width=0.5\textwidth]{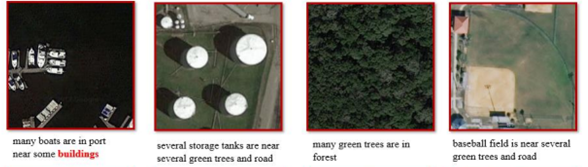}
    \caption{Image to Caption Results on an xView image}
    \label{caption}
\end{figure}

The following image and text description are from the NASNetMobile \cite{zoph2018learning} model on random xView \cite{lam2018xview} images [Figure \ref{caption}]. These previously uncaptioned images could then be sent to an ally unit or to a database, where searching images with text \cite{shivamshrirao} becomes possible. This image representation technique reduces the size of sent information from a picture to a sentence allowing reception of information even in low connectivity areas. BLEU \cite{papineni2002bleu} scores between prediction and the test set show NASNetMobile \cite{zoph2018learning} (0.650177, 0.474976, 0.405702, 0.297666 BLEU 1-4) as good enough for such a small (5.3M parameters) model.

\section{Conclusion}
While the speed of information is impacted by the increased processing time of the quality of information, cloud detection, super resolution, image to caption have remarkable potential to change the way we process satellite imagery along the sensor to shooter timeline. 

\section*{Acknowledgment}
The authors would like to thank the PeopleTec Technical Fellows program for encouragement and project assistance. The views and conclusions contained in this paper are those of the authors and should not be interpreted as representing any funding agencies.

\bibliographystyle{./IEEEtran.bst}
\bibliography{./refs.bib}

\end{document}